\documentclass[preprint]{acmart}

\usepackage{graphicx}
\usepackage{epstopdf}
\usepackage{booktabs} 
\usepackage{textcomp}
\usepackage{xcolor,colortbl}
\definecolor{White}{rgb}{1,1,1}
\definecolor{Gray}{gray}{0.9}
\definecolor{LightCyan}{rgb}{0.88,1,1}

\setcopyright{rightsretained}

\copyrightyear{2018}
\acmYear{2018}
\setcopyright{acmlicensed}
\acmConference[GECCO '18]{Genetic and Evolutionary Computation Conference}{July 15--19, 2018}{Kyoto, Japan}
\acmBooktitle{GECCO '18: Genetic and Evolutionary Computation Conference, July 15--19, 2018, Kyoto, Japan}
\acmPrice{15.00}
\acmDOI{10.1145/3205455.3205539}
\acmISBN{978-1-4503-5618-3/18/07}

\begin{document}
\title{Where are we now? \\A large benchmark study of recent symbolic regression methods}

\author{Patryk Orzechowski}
\affiliation{%
  \institution{University of Pennsylvania}
  \streetaddress{3700 Hamilton Walk}
  \city{Philadelphia, PA 19104, USA}
}
\email{patryk.orzechowski@gmail.com}

\author{William La~Cava}
\authornote{corresponding author}
\affiliation{%
  \institution{University of Pennsylvania}
  \streetaddress{3700 Hamilton Walk}
  \city{Philadelphia, PA 19104, USA}
}
\email{lacava@upenn.edu}

\author{Jason H. Moore}
\affiliation{%
  \institution{University of Pennsylvania}
  \streetaddress{3700 Hamilton Walk}
  \city{Philadelphia, PA 19104}
}
\email{jhmoore@upenn.edu}

\renewcommand{\shortauthors}{P. Orzechowski et al.}

\begin{abstract}
In this paper we provide a broad benchmarking of recent genetic programming approaches to symbolic regression in the context of state of the art machine learning approaches. We use a set of nearly 100 regression benchmark problems culled from open source repositories across the web. We conduct a rigorous benchmarking of four recent symbolic regression approaches as well as nine machine learning approaches from scikit-learn. The results suggest that symbolic regression performs strongly compared to state-of-the-art gradient boosting algorithms, although in terms of running times is among the slowest of the available methodologies. We discuss the results in detail and point to future research directions that may allow symbolic regression to gain wider adoption in the machine learning community.
\end{abstract}

%
%



\maketitle

\section{Introduction}
Since the beginning of the field, the genetic programming (GP) community has considered the task of symbolic regression (SR) as a basis for methodology research and as a primary application area. GP-based SR (GPSR) has produced a number of notable results in real-world regression applications, for example dynamical system modeling in physics~\cite{schmidt_distilling_2009}, biology~\cite{schmidt_automated_2011}, industrial wind turbines~\cite{la_cava_automatic_2016}, fluid dynamics~\cite{la_cava_inference_2016}, robotics~\cite{bongard_nonlinear_2005}, climate change forecasting~\cite{stanislawska_modeling_2012}, and financial trading~\cite{korns_accuracy_2011}, among others. However, the most prevalent use of GPSR is in the experimental analysis of new methods, for which SR provides a convenient platform for benchmarking. Despite this persistent use, several shortcomings of SR benchmarking are notable. First, the GP community lacks a unified standard for SR benchmark \textit{datasets}, as noted previously~\cite{mcdermott_genetic_2012}. Several SR benchmarks have been proposed \cite{nguyen_subtree_2015,korns_accuracy_2011,vladislavleva_order_2009}, critiqued \cite{dick_re-examination_2015,mcdermott_genetic_2012}, and black-listed~\cite{mcdermott_genetic_2012}, leading to inconsistencies in the experimental design of papers. In addition to a lack of consensus for benchmark datasets, there is not a standard set of benchmark \textit{algorithms} against which new methods are compared. As a result, it is typical for researchers to design or choose their own set of algorithms to compare to proposed methods, and it is up to reviewers and readers to assess the validity of the comparison. Experiments typically consider single values for GP hyperparameters such as population size or crossover rate, which increases the uncertainty of results even further. These practices make it nearly impossible to judge a new method outside the narrow scope of the experimental results.

Of course, there are shortcomings to focusing on benchmarks as well, as noted by others~\cite{white_better_2012,drummond_warning:_2010}. Putting too much focus on benchmarking may stifle innovation or lead to a lack of generalization to new tasks. However, the evidence suggests that the GP community is far from being overly focused on benchmarking. A 2012 survey of GP papers in EuroGP and GECCO from 2009 - 2011 reported the average number of SR problems per paper to be 2.4~\cite{mcdermott_genetic_2012}; 26.2\% of papers relied on the quartic polynomial problem, which has since been black-listed for being too trivial~\cite{white_better_2012}. We contend that the lack of focus in the GP community on rigorous benchmarking makes it hard to know how GPSR methods fit into the broader machine learning (ML) community. This lack of clarity also impedes the adoption of advancements to traditional GP techniques, and leaves researchers unsure about which advancements will have meaningful impacts.

There have been a few efforts to conduct broad benchmarking of GP methods in the past. For example, a recent a study looked at five SR methods on a set of five synthetic and four real world datasets~\cite{zegklitz_symbolic_2017}. Outside of GP, the efforts to benchmark ML approaches across many problems are more frequent, although most focus on the task of classification. Previous studies have looked at hundreds classification methodologies~\cite{fernandez-delgado_we_2014} and up to 165 datasets~\cite{olson_data-driven_2017}. Collaborative online tools such as Kaggle and OpenML~\cite{vanschoren_openml_2014} have also driven ML benchmarking and adoption of new methods. These larger benchmark studies have, for the most part, ignored GP-based methods. As a result, the GPSR field lacks a general sense of where it stands in relation to the broader ML field in terms of expected performance.

Our goal in this study is to present initial results in our efforts to assess the performance of recent GPSR methods in the broad context of ML regression. We benchmark the performance of four recent SR algorithms and ten established ML approaches on a collection of 94 different real-world regression problems. For each problem we consider hyperparameter tuning via cross-validation and assess each method in terms of training error, test error, and wall-clock time. Finally, we provide the code for the analysis in order to allow researchers to benchmark their own methods in this framework and reproduce the results shown here. 

\section{Methods}
We compare four recent GPSR methods in this benchmark and ten well-established ML regression methods. In this section we briefly present the selected methods and describe the design of the experiment.

\subsection{GP methods}
A number of factors impacted our choice of these methods. Two key elements were open-source implementations and ease of use. In addition, we wished to test different research thrusts in GP literature. The four methods encompass different innovations to standard GPSR, including incorporation of constant optimization, semantic search divers, and Pareto optimization. Each method is described briefly below. 

\paragraph{Multiple regression genetic programming (MRGP)}~\cite{Arnaldo2014} MRGP combines Lasso regression with the tree search afforded by GP. A weight is attached to each node in each program. These weights are adapted by applying Lasso regression to the entire program trace. MRGP uses point mutation and sub-tree crossover for variation and NSGA-II for selection. We use the version implemented in FlexGP~\footnote{https://flexgp.github.io/gp-learners/}.
\paragraph{$\epsilon$-Lexicase selection (EPLEX)}~\cite{la_cava_epsilon-lexicase_2016} $\epsilon$-lexicase selection adapts lexicase selection method~\cite{spector_assessment_2012} for regression. Rather than aggregating performance on the training set into a single fitness score, EPLEX selects parents by filtering the population through randomized orderings of training samples and removing individuals that are not within $\epsilon$ of the best performance in the pool. We use the EPLEX method implemented in ellyn\footnote{https://epistasislab.github.io/ellyn/}. Ellyn is a stack-based GP system written in C++ with a Python interface for use with scikit-learn. It uses point mutation and subtree crossover. Weights in the programs are trained each generation via stochastic hill climbing. A Pareto archive of trade-offs between mean squared error and complexity is kept during each run, and a small internal validation fold is used to select the final model returned by the search process. 
\paragraph{Age-fitness Pareto Optimization (AFP)}~\cite{schmidt_age-fitness_2011} AFP is a selection scheme based on the concept of age-layered populations introduced by Hornby et. al. ~\cite{hornby_alps2006}. AFP introduces a new individual each generation with an age of 0. An individual's age is updated each generation to reflect the number of generations since its oldest node (gene) entered the population. Parent selection is random and Pareto tournaments are used for survival on the basis of age and fitness. We use the version of AFP implemented in ellyn, with the same settings described above. 
\paragraph{Geometric Semantic Genetic Programming (GSGP)}~\cite{moraglio_geometric_2012} GSGP is a recent method that has shown many promising results for SR and other tasks. The main concept behind GSGP is the use of semantic variation operators that produce offspring whose semantics lie on the vector between the semantics of the parent and the target semantics (i.e. target labels). Use of these variation operators has the advantage of creating a unimodal fitness landscape. On the downside, the variation operators result in exponential growth of programs. We use the version GSGP implemented in C++ by Castelli et. al. \cite{castelli_c++_2015}, which is optimized to minimize memory usage. It is available from SourceForge\footnote{http://gsgp.sourceforge.net/}.

\subsection{ML methods}
We use scikit-learn~\cite{pedregosa2011} implementations of the following methods in this study:

\paragraph{Linear Regression} Linear Regression is a simple model of regression that minimizes the sum of the square errors of a linear model of inputs. The model is defined by $\hat{y}=b+w^T x$, where $y$ is a dependent variable (target), $x$ are explanatory variables, $b$ and $w$ are intercept and slope variables, and the minimized function is equal to (1).

\begin{equation}
C_{LR}(w)=\frac{1}{2} \sum _i{(y_i - w^T x_i)^2}
\end{equation}

\paragraph{Kernel Ridge} Kernel Ridge \cite{Robert2014} performs Ridge regression using a linear function in the space of the respective kernel. Least squares with l2-norm regularization is applied in order to prevent overfitting. The minimized function is equal to (2), where $\phi$ is a kernel function and $\lambda$ is the regularization parameter.
\begin{equation}
C_{KR}(w)=\frac{1}{2} \sum _i{(y_i - w^T \phi(x_i))^2}+\frac{1}{2} \lambda||w||^2
\end{equation}

\paragraph{Least-angle regression with Lasso} Lasso (Least absolute shrinkage and selection operator) is a popular method of regression that applies both feature selection and regularization \cite{tibshirani1996regression}. Similarly to Kernel Ridge, high values of $w$ are penalized. The use of the l1-norm on $w$ in the minimization function (see (3)) improves the ability to push individual weights to zero, effectively performing feature selection.
\begin{equation}
C_{L}(w)=\frac{1}{2} \sum _i{(y_i - w^T \phi(x_i))^2} + \lambda ||w||_1
\end{equation}
Least-angle regression with Lasso, a.k.a. Lars \cite{Efron2004}, is an efficient algorithm for producing a family of Lasso solutions. It is able to compute the exact values of $\lambda$ for new variables entering the model.

\paragraph{Linear SVR} Linear Support Vector Regression extends the concept of Support Vector Classifiers (SVC) to the task of regression, i.e. to predict real values instead of classes. Its objective is to minimize an $\epsilon$-insensitive loss function with a regularization penalty ($\frac{1}{2}||w||^2$) in order to improve generalization \cite{smola2004tutorial}. 

\paragraph{SGD Regression} SGD Regression implements stochastic gradient descent and is especially well suited for larger problems with over 10,000 of instances \cite{pedregosa2011}. We add this method of regression regardless, to compare its performance on smaller datasets.

\paragraph{MLP Regressor} Neural networks have been applied to regression problems for almost three decades \cite{hinton1989connectionist}. We include multilayer perceptrons (MLPs) as one of the benchmarked algorithms. We decided to benchmark neural network with a single hidden layer with fixed number of neurons (100) and compare different activation functions, learning functions and solvers, including the novel adam solver~\cite{kingma2014adam}. 

\paragraph{AdaBoost regression} Adaptive Boosting, called also AdaBoost \cite{Freund1997,Drucker1997}, is a flexible technique of combining a set of weak learners into a single stronger regressor. By changing the distribution (i.e. weights) of instances in the data, previously misclassified instances are favored in consecutive iterations. The final prediction is obtained by a weighted sum or weighted majority voting. As the result, the final regressor has smaller prediction errors. The method is considered sensitive to outliers.

\paragraph{Random Forest regression} Random Forests \cite{Breiman2001} are a very popular ensemble method based on combining multiple decision trees into a single stronger predictor. Each tree is trained independently with a randomly selected subset of the instances, in a process known as bootstrap-aggregating or bagging. The resulting prediction is an average of multiple predictions. Random forests try to reduce variance by not allowing decision trees to grow large, making them harder to overfit. 

\paragraph{Gradient Boosting regression} Gradient Boosting \cite{Friedman2001} is an ensemble method that is based on regression trees. It shares the AdaBoost concept of iteratively improving the system performance on its weakest points. In contrast to AdaBoost, the distribution of the samples remain the same. Instead, consecutively created trees correct the errors of the previous ones. Gradient Boosting minimizes bias (not variance like in Random Forests). In comparison to Random Forests, Gradient Boosting is sequential (thus slower), more difficult to train, but is reported to perform better than Random Forest~\cite{olson_data-driven_2017}.

\paragraph{Extreme Gradient Boosting} Extreme Gradient Boosting, also known as XGBoost \cite{Chen2016}, incorporates regularization into the Gradient Boosting algorithm in order to control overfitting. Its objective function combines the optimization of training loss with model complexity. This brings the predictor closer to the underlying distribution of the data, while encouraging simple models, which have smaller variance. Extreme gradient boosting is considered a state-of-the-art method in ML.

\subsection{Datasets}
\begin{figure}[ht!]
\begin{center}
\includegraphics[width=0.45\textwidth]{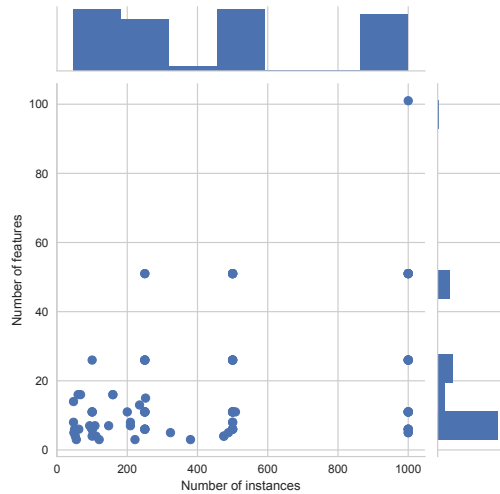}
\end{center}
\caption[Datasets]{Basic characteristics of the datasets used in the study}
\label{datasets}
\end{figure}

We pulled the benchmark datasets from the Penn Machine Learning Benchmark (PMLB) \cite{olson_pmlb:_2017} repository\footnote{https://github.com/EpistasisLab/penn-ml-benchmarks}, which contains a large collection of standardized datasets for classification and regression problems. This repository overlaps heavily with datasets from UCI, OpenML, and Kaggle. In this paper we considered regression problems only, of which there are 120 total. For our experiment, we removed datasets with 3000 instances or more (22 datasets) and two others for which at least one of the methods failed to provide the required number of results in feasible time (i.e. 72 hours on Intel\textregistered~ Xeon\textregistered ~ CPU E5-2680 v3 @ 2.50GHz). This gave the collection of 94 datasets in total. The distribution of the number of instances and the number of features in the collection of the datasets is presented in Fig. \ref{datasets}.

\subsection{Experiment design}

\begin{table*}[ht]
\caption{Analyzed algorithms with their parameters settings. The parameters in quotations refer to their names in the scikit-learn implementations.}
\label{settings}
\footnotesize
\begin{center}
\begin{tabular}{c c c}
\textbf{Algorithm name} &\textbf{Parameter} &\textbf{Values}  \\
\toprule
eplex, & pop size / generations & \{100/1000,1000/100\} \\
afp,        & max program length / max depth & \{64 / 6\} \\
mrgp       & crossover rate & \{0.2,0.5,0.8\} \\
        & mutation rate & 1-crossover rate \\
\midrule
gsgp & pop size / generations & \{100/1000,200/500,1000/100\} \\
        & initial depth & \{6\} \\
       & crossover rate & \{0.0,0.1,0.2\} \\
        & mutation rate & 1-crossover rate \\
 \midrule
eplex\_1M & pop size / generations& \{500/2000,1000/1000,2000/500\} \\
        &max program length & \{100\} \\
        &crossover rate & \{0.2,0.5,0.8\} \\
        &mutation rate & 1-crossover rate \\
\midrule
AdaBoostRegressor & `n\_estimators' & \{10, 100, 1000\} \\ 
        &`learning\_rate' & \{0.01, 0.1, 1, 10\} \\
\midrule
GradientBoostingRegressor & `n\_estimators' & \{10, 100, 1000\} \\
    &`min\_weight\_fraction\_leaf' & \{0.0, 0.25, 0.5\} \\
    &`max\_features'& \{`sqrt',`log2', None\} \\
\midrule
KernelRidge 
    &`kernel'& \{`linear', `poly', `rbf', `sigmoid'\} \\ 
    &`alpha'& \{1e-4, 1e-2, 0.1, 1\} \\
    &`gamma'& \{0.01, 0.1, 1, 10 \} \\
\midrule
LassoLARS & `alpha' & \{ 1e-04, 0.001, 0.01, 0.1, 1 \} \\
\midrule
LinearRegression &default &default\\
\midrule
MLPRegressor &`activation' & \{`logistic', `tanh', `relu'\} \\
        &`solver' &\{`lbfgs','adam',`sgd'\} \\
        &`learning\_rate' & \{`constant', `invscaling', `adaptive'\} \\
\midrule
RandomForestRegressor & `n\_estimators' & \{10, 100, 1000\} \\
    &`min\_weight\_fraction\_leaf' & \{0.0, 0.25, 0.5\} \\
    &`max\_features'& \{`'sqrt',`log2', None\} \\

\midrule
SGDRegressor &`alpha'& \{1e-06, 1e-04, 0.01, 1 \} \\
        &`penalty'& \{`l2', `l1', `elasticnet'\} \\
\midrule
LinearSVR &`C' & \{1e-06, 1e-04, 0.1, 1 \} \\
        &`loss' & \{`epsilon\_insensitive', `squared\_epsilon\_insensitive'\} \\
\midrule
XGBoost &`n\_estimators' & \{10, 50, 100, 250, 500, 1000\} \\ 
        &`learning\_rate' & \{1e-4, 0.01, 0.05, 0.1, 0.2\} \\
        &`gamma' & \{0, 0.1, 0.2, 0.3, 0.4\}\\
        &`max\_depth' & \{6\} \\
        &`subsample' & \{0.5, 0.75, 1\} \\
\midrule
\end{tabular}
\end{center}
\end{table*}

In order to benchmark different regression methods, an effort was made to measure performance of each of the methods in as similar an environment as possible. First, we decided to treat each of the GP methods as a classical ML approach and used the scikit-learn library \cite{pedregosa2011} for cross validation and hyperparameter optimization. This required some source code modifications to allow GSGP and MRGP to communicate with the wrapper.  Second, instead of reimplementing the algorithms, we relied on the original implementations with as few modifications as possible. Wrapping each method allowed us to keep a common benchmarking framework based on the scikit-learn functions.

The datasets were divided in the following way: 75\% of samples in each of the datasets were used for training, whereas the remaining 25\% were used for testing. We used grid search to tune the hyperparameters of each method.  Each method was run with a preset grid of input parameters, detailed in Table~\ref{settings}. The optimal setting of the parameters was determined based on 5-fold cross-validation performed on training data only. The setting with the best $R^2$ score across all folds was used for training the algorithms on the whole training data. The performance of the methods was measured on both training and testing datasets on the best model obtained during cross-validation. We repeated the entire experiment 10 times for each method and dataset.

Because of time constraints, we decided to run each of the GP-based methods for 100,000 evaluations (population size x number of generations). Additionally, we generated results for 1 million evaluations using EPLEX (referred to as EPLEX\_1M) in order to assess how much a more thorough training of a GP-based regressor would improve its performance.
\paragraph{Data preprocessing}
We decided to feed benchmarked algorithms with scaled data using StandardScaler function from scikit-learn. The reason for this is our effort to keep the format of the input data consistent across multiple algorithms for the purpose of benchmarking. The choice of the optimal preprocessing method for the particular regressor is out of scope of this paper.

\paragraph{Initialization of the algorithms}
We initially considered starting each of the methods with the same random seed, but eventually decided to make all data splits randomly. In our belief both approaches have disadvantages: the results will either be biased by the choice of the random seed, or by using different splits for different methods. By taking a median of the scores we became independent of the initial split of the data.

\paragraph{Wrappers for the GP methods.}
Some modifications needed to be done to each of the GP methods.
For EPLEX and AFP, ellyn provides an existing Python wrapper that was used. For other methods we implemented a class derived from scikit-learn \emph{BaseEstimator}, which implemented two methods: \emph{fit()}, used for training the regressor, and \emph{predict()}, used for testing performance of the regressor. The source code of MRGP and GSGP had to be modified, so that the algorithms could communicate with the wrapper. 

\paragraph{Parameters for the algorithms.}
The settings of the input parameters for the algorithms were determined based on the available recommendations for the given method, as well as previous experience of the authors. For GP-based methods we applied from 6 to 9 different settings (mainly: population size x number of generations and crossover and mutation rates). For the ML algorithms the number of settings was method dependent. The largest grid of the parameters was used for XGBoost method. The exact parameters for the methods used in this study can be found in Table \ref{settings}. 

\begin{figure*}
\begin{center}
\includegraphics[width=\textwidth]{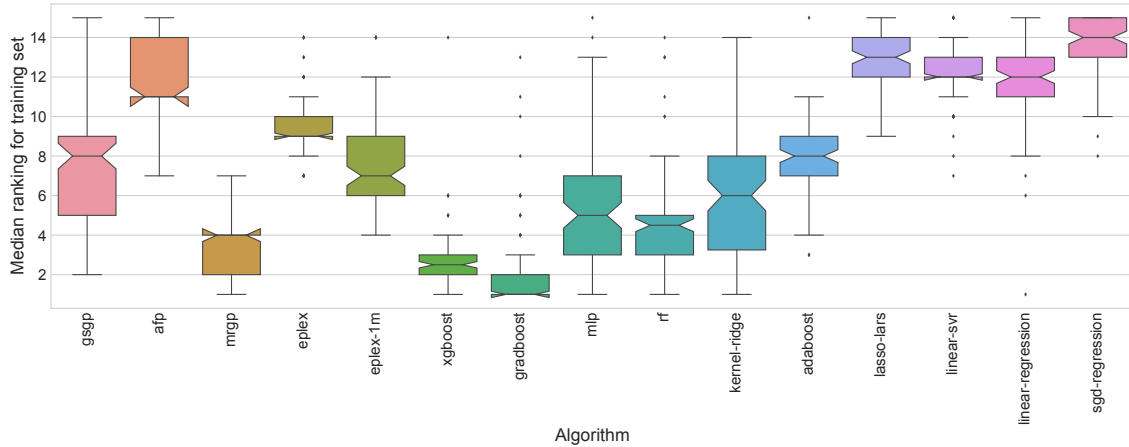}
\end{center}
\caption{Ranking of the performance of the algorithms based on the MSE score on training datasets.}
\label{training}
\end{figure*}

\begin{figure*}
\begin{center}
\includegraphics[width=\textwidth]{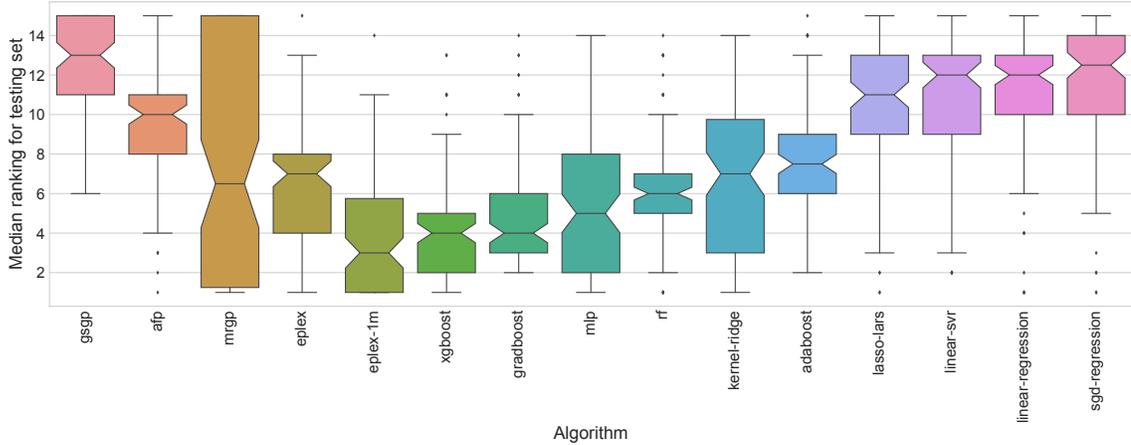}
\end{center}
\caption{Ranking of the performance of the algorithms based on the MSE score on testing datasets.}
\label{testing}
\end{figure*}

\section{Results}

We present aggregated results of the benchmarked algorithms on the collection of 94 regression datasets in Figures~\ref{training}-\ref{testing}. The relative performance of the algorithms was determined as the ability to make the best predictions on the training and testing data using mean squared error (MSE) of the samples. The performance on the testing dataset is of primary importance, as it shows how well the methods can generalize to previously unseen data \cite{Domingos2012}. However we include the training comparisons as a way to assess the prediliction for overfitting among methods.

We first analyze the results for each of the regression tasks on the training data. The relative rankings of each method in terms of MSE is presented in Fig. \ref{training}. The best training performance was obtained with gradient boosting, which completed in top-2 for the vast majority of the benchmarked datasets. The second best method across all the datasets was XGBoost. The top-performing GP method across all the datasets was MRGP and held the third place on average across training sets.

The test data results allow us to assess how well the algorithms handle generalization as well as their level of overfitting on training data. The relative performance of the methods changed noticeably when previously unseen data was used for evaluation. The results are presented in Fig. \ref{testing}. The best performing method on average was EPLEX-1M. This GPSR method slightly outperformed XGBoost, which ended as the second best generalizing method across datasets. Gradient boosting was the third best method, and MLP finished in fourth place. 

Several of the methods exhibit overfitting by changing ranking between the training and testing evaluations. Gradient boosting, for example, moves from first to third place. The performance of MRGP, which was one of the best regressors on the training data, also exhibits overfitting, resulting in a drop of its average ranking from 4th to 6th.  MRGP's results also contained the highest variance in performance on test sets. GSGP exhibits the highest level of overfitting in terms rank changes, dropping from 8th to 13th. Conversely, several methods appear to generalize well, including EPLEX-1M (moving from a median ranking of 5 to 3) and Lasso (13 to 11). 

We used the test set MSE scores to check for significant differences between methods across all datasets according to a Friedman test, which produces a $p$-value less than 2e-16, indicating significant differences. Post-hoc pairwise tests are then conducted and reported in Table~\ref{tbl:stats}. The large number of datasets provides higher statistical power than smaller experimental studies, leading to many $p$-values below 0.05. EPLEX-1M statistically outperforms the highest number of other methods (11), followed by XGBoost (9) and gradient boosting (7). We find that none of the comparisons between EPLEX-1M, XGBoost, gradient boosting and MLP are significantly different.

We now analyze the GP methods given equivalent numbers of fitness evaluations (AFP, MRGP, EPLEX, and GSGP). The results between MRGP and EPLEX show no significant difference. The only noted difference is that EPLEX significantly outperforms AFP, whereas MRGP does not. The three methods AFP, MRGP, and EPLEX all significantly outperform GSGP. Given more fitness evaluations, EPLEX-1M significantly outperforms all the other GP experiments, including EPLEX.  

\begin{table*}
\scriptsize
\caption{Friedman Asymptotic General Symmetry Test. Bold indicates $p<$0.05.}\label{tbl:stats}

\begin{tabular}{l r r r r r r r r r r r r r r} \toprule

& eplex-1m & xgboost & gradboost & mlp & rf & eplex & mrgp & kernel- & adaboost & afp & lasso- & linear- & linear- 		& sgd- \\ 
&		   &		 &			 &	   &	&	    &	   & ridge &          &       & lars & svr   & regression & regression 	\\ \midrule
xgboost	&  { 1 }& -& -& -& -& -& -& -& -& -& -& -& -& -\\
\rowcolor{LightCyan}
gradboost	&  { 0.9 }&  { 1 }& -& -& -& -& -& -& -& -& -& -& -& -\\
mlp	&  { 0.2 }&  { 0.6 }&  { 1 }& -& -& -& -& -& -& -& -& -& -& -\\
\rowcolor{LightCyan}
rf	& \textbf { 0.003 }&  { 0.05 }&  { 0.6 }&  { 1 }& -& -& -& -& -& -& -& -& -& -\\
eplex	& \textbf { 0.001 }& \textbf { 0.02 }&  { 0.4 }&  { 1 }&  { 1 }& -& -& -& -& -& -& -& -& -\\
\rowcolor{LightCyan}
mrgp	& \textbf { 3e-07 }& \textbf { 2e-05 }& \textbf { 0.005 }&  { 0.2 }&  { 0.9 }&  { 1 }& -& -& -& -& -& -& -& -\\
kernel-ridge	& \textbf { 0.0007 }& \textbf { 0.02 }&  { 0.4 }&  { 1 }&  { 1 }&  { 1 }&  { 1 }& -& -& -& -& -& -& -\\
\rowcolor{LightCyan}
adaboost	& \textbf { 1e-07 }& \textbf { 4e-06 }& \textbf { 0.002 }&  { 0.1 }&  { 0.8 }&  { 0.9 }&  { 1 }&  { 0.9 }& -& -& -& -& -& -\\
afp	& \textbf { 3e-16 }& \textbf { 7e-14 }& \textbf { 4e-10 }& \textbf { 9e-06 }& \textbf { 0.0008 }& \textbf { 0.002 }&  { 0.3 }& \textbf { 0.004 }&  { 0.5 }& -& -& -& -& -\\
\rowcolor{LightCyan}
lasso-lars	& \textbf { 0 }& \textbf { 0 }& \textbf { 2e-15 }& \textbf { 1e-11 }& \textbf { 1e-07 }& \textbf { 5e-07 }& \textbf { 0.002 }& \textbf { 6e-07 }& \textbf { 0.006 }&  { 1 }& -& -& -& -\\
linear-svr	& \textbf { 0 }& \textbf { 0 }& \textbf { 0 }& \textbf { 3e-13 }& \textbf { 2e-09 }& \textbf { 3e-08 }& \textbf { 0.0002 }& \textbf { 7e-08 }& \textbf { 0.0008 }&  { 0.8 }&  { 1 }& -& -& -\\
\rowcolor{LightCyan}
linear-regression	& \textbf { 0 }& \textbf { 0 }& \textbf { 0 }& \textbf { 1e-14 }& \textbf { 7e-11 }& \textbf { 6e-10 }& \textbf { 5e-05 }& \textbf { 1e-09 }& \textbf { 0.0001 }&  { 0.5 }&  { 1 }&  { 1 }& -& -\\
sgd-regression	& \textbf { 0 }& \textbf { 0 }& \textbf { 0 }& \textbf { 0 }& \textbf { 1e-13 }& \textbf { 4e-12 }& \textbf { 1e-07 }& \textbf { 1e-12 }& \textbf { 5e-07 }&  { 0.07 }&  { 0.9 }&  { 1 }&  { 1 }& -\\
\rowcolor{LightCyan}
gsgp	& \textbf { 0 }& \textbf { 0 }& \textbf { 0 }& \textbf { 0 }& \textbf { 0 }& \textbf { 0 }& \textbf { 2e-12 }& \textbf { 0 }& \textbf { 2e-11 }& \textbf { 0.0004 }&  { 0.1 }&  { 0.4 }&  { 0.7 }&  { 1 }\\
\bottomrule

\end{tabular}
\end{table*}

\begin{figure*}[ht!]
\begin{center}
\includegraphics[width=\textwidth]{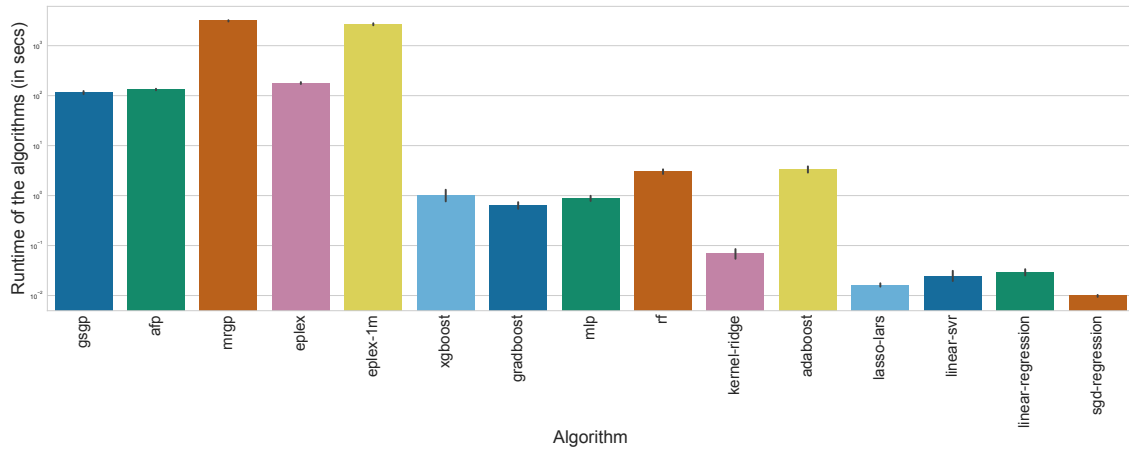}
\end{center}
\caption{Median running time of each of the algorithms (in seconds).}
\label{runtime}
\end{figure*}

The comparison of running times per training task is presented in Fig. \ref{runtime}. Three important considerations should be made when assessing these results. First, the experiment was conducted in a cluster environment. Second, each algorithm was run on a single thread for each dataset. Thus the easily parallelized algorithms (i.e., all GP-based methods and some ensemble tree methods) would likely show better relative performance in a multicore setting. 
Third, benchmarked algorithms were implemented using different programming languages. Thus, comparison of running times doesn't exclusively reflect the complexity of the methods.

Despite these considerations, it is worth noting how much additional computation time is required by the GP methods, which are one to three orders of magnitude slower than the nearest comparison. In terms of GP methods, MRGP runs the slowest, which may be partially due to its Java implementation (the other four GP methods use c++). EPLEX-1M is able to complete 10 times as many fitness evaluations in approximately the same time. The other three GP methods (GSGP, AFP, and EPLEX) show similar computation times. Among other ML methods, the ensemble tree methods and MLP are the slowest, and the linear methods are fastest, as expected. 

The most frequent settings of the parameters picked for the best model across all trials are presented in Table \ref{best-settings}. We purposefully do not include Linear Regression in the table (run with the default values) or Kernel Ridge regression, for which multiple settings of input parameters performed comparably. It may be noted each GP-based method besides GSGP tended to favor large population sizes over larger numbers of generations. The optimal setting for crossover and mutation rates varied beteween methods.

\begin{table*}[ht]
\caption{Most frequently chosen parameter settings based on 5-fold cross validation across all datasets.}\label{best-settings}
\footnotesize
\begin{center}
\def\arraystretch{1.5}%
\begin{tabular}{c c }
\textbf{Algorithm name} &\textbf{Frequently best parameter settings} \\
\hline
    gsgp &(`g'=500, `max\_len'=6, `popsize'=200, `rt\_cross'=0.2, `rt\_mut'=0.8) \\   
    afp & (`g'=100, `max\_len'=64, `popsize'=1000, `rt\_cross'=0.8, `rt\_mut' 0.2) \\
    mrgp & (\{`g'=100, `pop\_size'=1000\} or the opposite; 'rt\_cross'=0.2, `rt\_mut'=0.8) \\
    eplex & (`g'=100, `max\_len'=64, `popsize'=1000, `rt\_cross'=0.8, `rt\_mut'=0.2)\\
    eplex-1m & (`g'=500, `max\_len'=100, `popsize'=2000, `rt\_cross'=0.8, `rt\_mut'=0.2)\\   
    xgboost & (`'gamma'=0, `learning\_rate'=0.01, `max\_depth'=6, `n\_estimators'=1000, 'subsample'=0.5) \\    
	gradboost & (`max\_features'=None, `min\_weight\_fraction\_leaf'=0.0, `n\_estimators'=1000) \\
    mlp & (`activation'='logistic', `learning\_rate'= `constant',`solver'=`lbfgs') \\   
    rf & (`max\_features'=None, `min\_weight\_fraction\_leaf'=0.0, `n\_estimators'=1000) \\
    adaboost &(`learning\_rate'=1.0, `n\_estimators'=1000) \\    
    lasso-lars & (`alpha'=`0.001') \\
    linear-svr&  (`C'=0.1, `loss'=`squared\_epsilon\_insensitive') \\    
 	sgd-regression & (`alpha'=0.01, `penalty'=`l1') \\
 	linear-regression & (`fit\_intercept' `True') \\
    \hline    
\end{tabular}
\end{center}
\end{table*}


\section{Conclusions}

In this paper we evaluated four recent GPSR methods in comparison to ten state-of-the-art ML methods on a set of 94 real-world regression problems. We consider hyper-parameter optimization for each method using nested cross-validation, and compare the methods in terms of the MSE they produce on training and testing sets, and their runtime. The analysis includes some interesting results. The most noteworthy finding is that a GPSR method ($\epsilon$-lexicase selection implemented in ellyn), given 1 million fitness evaluations, achieves the best test set MSE ranking across all datasets and methods. 
Two of the GP-based methods, namely: EPLEX and MRGP, produce competitive results compared to state-of-the-art ML regression approaches. The downside of the GP-based methods is their computation complexity when run on a single thread, which contributes to much higher runtimes. Parallelism is likely to be a key factor in allowing GP-based approaches to become competitive with leading ML methods with respect to running times. 

We also should note some shortcomings of this study that motivate further analysis. First, a guiding motivation for the use of GPSR is often its ability to produce legible symbolic models. Our analysis did not attempt to quantify the complexity of the models produced by any of the methods. An extension of this work could establish a standardized way of assessing this complexity, for example using the polynomial complexity method proposed by Vladislavleva et. al.~\cite{vladislavleva_order_2009}. Ultimately the relative value of explainability versus predictive power will depend on the application domain. Second, we have considered real world datasets for the source of our benchmarks. Simulation studies could also be used, and have the advantage of providing ground truth about the underlying process, as well as the ability to scale complexity or difficulty. It should also be noted that the datasets used for this study were of relatively small sizes (up to 1000 of instances). Future work should consider larger dataset sizes, but will come with a larger computational burden.

We have also limited our initial analysis to looking at bulk performance of algorithms over many datasets. Further analysis of these results should provide insight into the properties of datasets that make them amenable to, or difficult for, GP-based regression. Such an analysis can provide suggestions for new problem sub-types that may be of interest to the GP community. 

We hope this study will provide the ML community with a data-driven sense of how state-of-the-art SR methods compare broadly to other popular ML approaches to regression. 

\section*{Supplementary materials}
Source code for our experiment can be found at the following url: \url{https://github.com/EpistasisLab/regression-benchmark}.

\begin{acks}
The authors would like to thank Randal Olson, whose work helped motivate this study, and Weixuan Fu, who put together the regression benchmark resource. This research was supported in part by PL-Grid Infrastructure and by NIH grants LM010098 and AI116794.
\end{acks}

\bibliographystyle{ACM-Reference-Format}
\bibliography{sample-bibliography}

\end{document}